\documentclass[10pt,twocolumn,letterpaper]{article}

\usepackage{iccv}              

\definecolor{iccvblue}{rgb}{0.21,0.49,0.74}
\usepackage[pagebackref,breaklinks,colorlinks,allcolors=iccvblue]{hyperref}
\usepackage{amsmath,amssymb,amsfonts}
\usepackage{algorithmic}
\usepackage{graphicx}
\usepackage{textcomp}
\usepackage{multirow}
\usepackage{times}
\usepackage{epsfig}
\usepackage{bm}
\usepackage{graphicx}
\usepackage{colortbl}
\usepackage{booktabs}
\usepackage{subcaption}
\usepackage{amsmath}
\usepackage{amssymb}
\usepackage{pifont}
\newcommand{\cmark}{\ding{51}} 
\newcommand{\xmark}{\ding{55}} 
\usepackage{url}
\definecolor{tabhighlight}{HTML}{e5e5e5}

\newcommand{\tablestyle}[2]{\setlength{\tabcolsep}{#1}\renewcommand{\arraystretch}{#2}\centering\footnotesize}

\def\paperID{76} 
\def\confName{ICCV}
\def\confYear{2025}

\title{TextSAM-EUS: Text Prompt Learning for SAM to Accurately Segment Pancreatic Tumor in Endoscopic Ultrasound}

\author{
Pascal Spiegler$^{1\dagger}$, 
Taha Koleilat$^{1\dagger}$, 
Arash Harirpoush$^1$, 
Corey S. Miller$^{2,3,4}$, 
Hassan Rivaz$^1$, \\
Marta Kersten-Oertel$^1$, 
Yiming Xiao$^1$\\
$^1$Concordia University, Montreal, Canada\\
$^2$Jewish General Hospital, Montreal, Canada\\
$^3$McGill University Faculty of Medicine, Montreal, Canada\\
$^4$Lady Davis Institute for Medical Research, Montreal, Canada\\
{\tt\small pascal.spiegler@mail.concordia.ca}
\vspace{1mm} \\
$^{\dagger}$Equal contribution
}

\begin{document}
\maketitle

\begin{abstract}
Pancreatic cancer carries a poor prognosis and relies on endoscopic ultrasound (EUS) for targeted biopsy and radiotherapy. However, the speckle noise, low contrast, and unintuitive appearance of EUS make segmentation of pancreatic tumors with fully supervised deep learning (DL) models both error-prone and dependent on large, expert-curated annotation datasets. To address these challenges, we present \textsc{TextSAM-EUS}, a novel, lightweight, text-driven adaptation of the Segment Anything Model (SAM) that requires no manual geometric prompts at inference. Our approach leverages text prompt learning (context optimization) through the BiomedCLIP text encoder in conjunction with a LoRA-based adaptation of SAM’s architecture to enable automatic pancreatic tumor segmentation in EUS, tuning only 0.86\% of the total parameters. On the public Endoscopic Ultrasound Database of the Pancreas, \textsc{TextSAM-EUS} with automatic prompts attains 82.69\% Dice and 85.28\% normalized surface distance (NSD), and with manual geometric prompts reaches 83.10\% Dice and 85.70\% NSD, outperforming both existing state-of-the-art (SOTA) supervised DL models and foundation models (e.g., SAM and its variants). As the first attempt to incorporate prompt learning in SAM-based medical image segmentation, \textsc{TextSAM-EUS} offers a practical option for efficient and robust automatic EUS segmentation. Code is available at \url{https://github.com/HealthX-Lab/TextSAM-EUS}.
\end{abstract}    
\section{Introduction}
\label{sec:intro}

Pancreatic cancer is the sixth leading cause of cancer-related death, with a survival rate of approximately 10\% \cite{bray2024global}, highlighting its aggressive nature and an urgent demand for improved diagnostic and therapeutic strategies. Endoscopic ultrasound (EUS) is a medical imaging technique that involves inserting a thin, flexible tube equipped with an ultrasound probe into the digestive tract to obtain images of internal organs such as the pancreas. The procedure plays an important role in the clinical management of pancreatic cancer by enabling fine-needle aspiration for tissue sampling and the delivery of targeted cancer therapies (e.g., insertion of radioactive seeds) \cite{miller2024feasibility}. During such EUS procedures, accurate delineation of tumor boundaries can dramatically influence clinical decision-making, since precise probe placement and coverage are critical. However, the speckle noise, low contrast, and the unintuitive appearance of EUS images present challenges for automated segmentation. Fully supervised architectures, such as UNet-based models, have demonstrated strong performance in medical image segmentation, including tumor delineation, but often perform better on higher-contrast modalities (e.g., MRI and CT) than on noisier ultrasound. Moreover, these models require large amounts of pixel-wise annotations from many patients to perform well, limiting their practicality in clinical scenarios with limited labeled data, such as \textit{pancreatic tumor segmentation from EUS}. In contrast, the Segment Anything Model (SAM) \cite{kirillov2023segment}, a foundation model trained on over a billion images, supports zero-shot segmentation using manually-placed geometric prompts (e.g., points or boxes). However, in comparison to geometric prompts, text prompts could offer a more convenient alternative way to initiate segmentation without explicit judgment on tissue borders, since it can encode rich class-level information, but relevant explorations in this domain are limited. Additionally, SAM’s image encoder is pretrained entirely on natural images, leading to a significant domain shift when applied to medical images, particularly ultrasound images. Effective methods for parameter-efficient finetuning are therefore desirable to enable medical image application of these foundation models \cite{medsam}. 

To address these gaps, we propose \textsc{TextSAM-EUS}, a lightweight adaptation of SAM (tuning just 0.86\% of its parameters) tailored for pancreatic tumor segmentation in EUS. Our key contributions are as follows: \textbf{First}, we develop a novel framework that integrates text prompt learning (i.e., context optimization) to initiate EUS pancreatic tumor segmentation in a parameter-efficient low-rank adaptation (LoRA)-finetuned SAM model. \textbf{Second}, we introduce an iterative segmentation refinement step that supplements text prompts with geometric ones (bounding boxes and points) to enhance segmentation quality. \textbf{Third}, we benchmark \textsc{TextSAM-EUS} on a public pancreatic cancer EUS dataset, where it outperforms both UNet-based and other foundation models, including SAM-based methods, achieving 82.69\% Dice Similarity Coefficient (DSC) and 85.28\% Normalized Surface Distance (NSD) using fully automatic, text-driven inference.

\begin{table}[t]
    \caption{Comparison of trainable parameters and segmentation performance (DSC) between SAM ViT-B variants and the proposed \textsc{TextSAM-EUS}. \textsc{TextSAM-EUS} strikes the best balance between segmentation accuracy and parameter efficiency, achieving the highest DSC (82.69\%) with a minimal parameter footprint (1.69M), while requiring no expert manual intervention at inference.}
    \centering
    \tablestyle{-13pt}{1.1}
    \arrayrulecolor{black}
    \setlength\arrayrulewidth{1pt}
    \addtolength{\tabcolsep}{+16pt}
    \resizebox{\columnwidth}{!}{%
    \begin{tabular}{lccc}
    \toprule
    \multirow{2}{*}{Model} & Manual & \multirow{2}{*}{Params. (M) $\downarrow$} & \multirow{2}{*}{DSC (\%) $\uparrow$} \\
    & Prompts? & & \\
    \midrule
    SAM (Point)~\cite{kirillov2023segment} & \cmark & 93.74 & 39.80 \\
    SAM (Box)~\cite{kirillov2023segment} & \cmark & 93.74 & 78.15 \\
    SAMUS~\cite{lin2024samus} & \cmark & 39.65 & 70.75 \\
    MedSAM~\cite{medsam} & \cmark & 93.74 & 82.66 \\
    SAMed~\cite{samed} & \xmark & 1.45 & 78.00 \\
    AutoSAM~\cite{autosam} & \xmark & 41.56 & 81.26 \\
    AutoSAMUS~\cite{lin2024samus} & \xmark & 8.86 & 81.04 \\
    \rowcolor{gray!20} \textbf{TextSAM-EUS (Ours)} & \xmark & \textbf{1.69} & \textbf{82.69} \\
    \bottomrule
    \end{tabular}
    }
    \label{tab:params}
\end{table}

\section{Related Works}
\label{sec:related-work}
\subsection{SAM for Medical Imaging}
\label{subsec:SAM-for-medical-imaging}
The Segment Anything Model (SAM), a foundation model designed for promptable image segmentation, integrates a robust image encoder, a versatile prompt encoder, and a lightweight mask decoder to support zero-shot generalization. This architecture has sparked significant interest in the medical imaging community. For instance, MedSAM \cite{medsam} fine-tuned SAM on approximately one million medical image–mask pairs, achieving strong performance across a range of segmentation tasks. To improve efficiency, AutoSAM \cite{autosam} introduced an alternative fine-tuning strategy that involves training only the prompt encoder and using a specialized deconvolution-based decoder tailored for medical applications. Similarly, Zhang et al. \cite{samed} proposed SAMed, which attaches LoRA adapters to SAM’s image encoder and fine-tunes the prompt/mask decoders, enabling fully automatic medical image segmentation with minimal trainable parameters. Moreover, Wu et al. \cite{spfs} introduced SPFS-SAM, which equips SAM with a self-prompting mechanism where a lightweight classifier produces an initial mask from SAM’s embeddings and then automatically generates point prompts for iterative refinement. Additionally, Cheng \textit{et al.} \cite{Cheng2023SAMOM} systematically evaluated different prompt types and identified bounding boxes as the most effective for medical segmentation across 12 tasks. Addressing the issue of noisy pseudo-labels from SAM, Huang \textit{et al.} \cite{huang2023push} proposed a correction mechanism that refines these labels to improve downstream fine-tuning. In another direction, Gong \textit{et al.} \cite{gong20233dsamadapter} adapted the model to handle 3D volumetric medical data by substituting SAM's mask decoder with a 3D convolutional module. Furthermore, MedCLIP-SAM \cite{koleilat2024medclip, koleilat2024medclipv2} introduced the integration of vision-language pretraining into SAM by aligning the visual features from SAM with medical text embeddings from MedCLIP, enhancing zero-shot segmentation without requiring dense annotations. Furthermore, a recent uncertainty‐aware adaptation of SAM for intracranial hemorrhage segmentation \cite{spiegler2024weakly} combines YOLO‐based detection with an uncertainty‐rectified SAM framework to address weak supervision scenarios. Although these adaptations extend SAM's utility in the medical domain, they often depend on retraining procedures, prompt engineering, or manual geometric prompting.  In contrast, our method leverages textual prompts generated by BiomedCLIP, enabling manual prompt-free inference without requiring points, boxes, or masks.

\subsection{Ultrasound Pancreatic Tumor Segmentation}
Several prior works have explored the challenging task of segmenting pancreatic tumors in ultrasound (US) images. Traditional supervised methods, such as those proposed by Lu et al.~\cite{lu2021deep} and Huang et al.~\cite{huang2020cascade}, leverage modified U-Net architectures. Lu et al. introduced a multi-scale attention U-Net to focus on tumor regions in noisy ultrasound images, while Huang et al. employed a cascade segmentation framework with shape constraints to improve boundary delineation. Semi-supervised approaches have also been investigated, notably by Liu et al.~\cite{liu2021semi}, who proposed a multi-task consistency learning strategy to leverage both labeled and unlabeled data, addressing the scarcity of annotations in medical ultrasound datasets. More recently, foundation models have been adapted to ultrasound segmentation. SAMUS~\cite{lin2024samus} fine-tunes SAM with an additional CNN encoder, demonstrating promising results on US images. AutoSAMUS \cite{lin2024samus} builds on SAMUS by training a network to generate its point prompts automatically from image encoder features, removing the need for manual input. CC-SAM~\cite{ccsam2023} further improves the generalization of SAMUS through the use of grounding DINO to obtain geometric prompts through text. These efforts highlight a growing trend towards combining powerful foundation models with ultrasound-specific adaptations to enhance segmentation performance in low-contrast, noisy imaging modalities like ultrasound.

\subsection{VLMs for Biomedical Domain}
Vision-language models (VLMs) such as CLIP \cite{radford2021learning} and ALIGN \cite{jia2021scaling} have further advanced multimodal learning by employing contrastive self-supervision to map image and text representations into a joint embedding space. While these models excel at tasks like zero-shot classification and cross-modal retrieval in general domains, their performance often degrades in specialized fields like medicine, where subtle visual features and domain-specific language are crucial. To overcome these limitations, recent work has explored domain adaptation strategies, with prompt learning emerging as a computationally efficient alternative to full model fine-tuning. Approaches like CoOp \cite{zhou2022learning} and CoCoOp \cite{zhou2022conditional} perform text context optimization by learning task-specific text prompt tokens while keeping the base VLM frozen. Further developments, such as MaPLe \cite{khattak2023maple} and PromptSRC \cite{khattak2023self}, enhance generalization by introducing encoder tuning and regularization strategies. Adapter-based methods, including CLIP-Adapter \cite{gao2024clip} and Tip-Adapter \cite{zhang2021tip}, refine the visual encoder or leverage support sets for improved few-shot learning, though they may encounter training instability. In the biomedical space, several CLIP variants, such as BioViL \cite{boecking2022making}, PubMedCLIP \cite{eslami2021doesclipbenefitvisual}, and BiomedCLIP \cite{biomedclip}, have incorporated medical corpora to improve domain alignment. Other works \cite{koleilat2024medclip, koleilat2024medclipv2} further connect these general-purpose VLMs with clinical imaging applications. Nonetheless, capturing fine-grained clinical semantics remains a significant challenge \cite{xu2024advances, zhao2023clip}. Recent efforts have adapted CoOp-style prompting to the medical domain; DCPL \cite{cao2024domain} being a notable example, although these typically require moderate-to-large training sets. In contrast, BiomedCoOp \cite{koleilat2025biomedcoop} shows that strong performance can still be achieved in low-resource medical scenarios by leveraging text prompt tuning while maintaining generalization across tasks. While progress has been made in adapting CLIP and SAM individually for medical tasks, a unified end-to-end architecture that jointly leverages both models within the biomedical domain, particularly with efficient prompt learning, remains unexplored.
\section{Methods and Materials}
\label{sec:method}

A general overview of the proposed \textsc{TextSAM-EUS} framework is presented in Fig. \ref{fig:framework-fig}, which comprises learnable context tokens with the BiomedCLIP text encoder, LoRA adaptation of the SAM image encoder and mask decoder, as well as an iterative segmentation refinement module.

\subsection{SAM Preliminaries}
The Segment Anything Model \cite{kirillov2023segment} comprises three core components: a \textbf{large image encoder}, a \textbf{prompt encoder}, and a \textbf{lightweight mask decoder}.
\\
\\
\noindent \textbf{Image Encoder:} The image encoder $\bm{E}_{\text{img}}$ is a ViT-based backbone that processes an input image $\bm{X} \in \mathbb{R}^{3 \times H \times W}$ to produce a feature map:
\begin{equation}
\bm{F} = \bm{E}_{\text{img}}(\bm{X}) \in \mathbb{R}^{h \times w \times d}
\end{equation}
where $h \times w$ is the resolution of the encoded feature grid and $d$ is the feature dimension.
\\
\\
\noindent \textbf{Prompt Encoder:} The prompt encoder $\bm{E}_{\text{prompt}}$ maps user-defined prompts, such as points, bounding boxes, or masks, into prompt tokens:
\begin{equation}
\bm{P} = \bm{E}_{\text{prompt}}(\text{\textbf{prompt}}) \in \mathbb{R}^{k \times d}
\end{equation}
where $k$ depends on the number and type of prompts provided (e.g., number of points or box corners).
\\
\\
\noindent \textbf{Mask Decoder:} The decoder $\mathcal{D}_{\text{mask}}$ fuses image features $\bm{F}$ with prompt embeddings $\bm{P}$ to predict segmentation masks:
\begin{equation}
\hat{\bm{Y}} = \mathcal{D}_{\text{mask}}(\bm{F}, \bm{P}) \in \mathbb{R}^{H \times W}
\end{equation}
\noindent
This architecture enables SAM to generate segmentation masks conditioned on sparse or dense geometric prompts, supporting general-purpose segmentation across diverse domains.

\subsection{CLIP Preliminaries}
To enable text-driven segmentation, we leverage the CLIP \cite{radford2021learning} architecture (more specifically BiomedCLIP \cite{biomedclip}), which consists of a vision encoder $\bm{E_v}$ and a text encoder $\bm{E_t}$ that maps images and text into a shared embedding space. In our framework, we use the pre-trained BiomedCLIP text encoder $\bm{E_t}$ to encode class-specific textual prompts and feed them into the prompt encoder of SAM, effectively conditioning the segmentation process on natural language.

Given a batch of $B$ images and $C$ textual class prompts, the visual inputs are denoted as $\bm{X_v} \in \mathbb{R}^{B \times 3 \times H \times W}$, representing RGB images of size $H \times W$, and the text inputs as $\bm{X_t} \in \mathbb{R}^{C \times L}$, where $L$ is the tokenized sequence length. These are processed as:
\begin{equation}
\mathbf{T} = \bm{E_t}(\bm{X_t}) \in \mathbb{R}^{C \times D}
\end{equation}
where $\mathbf{T}$ contains the language embeddings used to guide the prompt encoder in SAM, and $D$ is the embedding dimension.

This approach enables SAM to be conditioned on text, allowing it to segment image regions based on natural language prompts. By incorporating BiomedCLIP's language priors into the SAM prompt encoder, our method enables text-driven segmentation capabilities that can benefit from better biomedical semantics.

\subsection{Low-rank Adaptation for SAM}
Although SAM demonstrates strong generalization on natural images, it is not pre-trained on medical images and fails to capture the subtle anatomical variations present in ultrasound scans \cite{lin2024samus}. In particular, the appearance of pancreatic tumors in US is highly variable and low in contrast, hindering robust zero-shot segmentation. To mitigate this domain gap, we apply LoRA \cite{hu2021lora} modules to both the SAM image encoder $\bm{E}_{\text{img}}$ and mask decoder $\mathcal{D}_{\text{mask}}$. Instead of updating the full weight matrices, LoRA introduces trainable low-rank updates to linear layers. For a weight matrix $\bm{W} \in \mathbb{R}^{d_{\text{out}} \times d_{\text{in}}}$, the adapted version becomes:
\begin{equation}
\bm{W}_{\text{LoRA}} = \bm{W} + \bm{A}\bm{B}, \quad \bm{A} \in \mathbb{R}^{d_{\text{out}} \times r}, \ \bm{B} \in \mathbb{R}^{r \times d_{\text{in}}},
\end{equation}
where $r \ll \min(d_{\text{in}}, d_{\text{out}})$ denotes the intrinsic rank of the adaptation, and \textbf{\textit{A }}and \textbf{\textit{B}} contain trainable parameters. This design allows efficient adaptation of SAM to ultrasound data while keeping the majority of the original model frozen. The LoRA modules are optimized jointly with the language-guided prompt tuning described next.

\subsection{Biomedical Text Prompt Learning}
To enable SAM to perform segmentation conditioned on biomedical text, we propose a CLIP-based text prompt learning strategy. Instead of using more commonly seen geometric prompts, we directly encode natural language queries upon context optimization \cite{khattak2024learning} using a BiomedCLIP \cite{biomedclip} encoder (PubMedBERT \cite{gu2021domain}), and then infuse this semantic representation into SAM's prompt encoder to initiate the segmentation.

Here, we introduce $b$ learnable prompt tokens $\{P^i \in \mathbb{R}^{d_t}\}_{i=1}^{b}$, where $d_t$ is the hidden dimension of the BiomedCLIP text encoder. These are concatenated with $S$ fixed tokens, denoted as $W_0 = [w^1, w^2, \cdots, w^S]$, such that the total number of tokens remains $N$, satisfying the fixed context length of the text encoder (i.e., $S + b = N$). The resulting sequence is passed through $K$ sequential Transformer layers:
\begin{align}
[P_0, \, W_0] &= [P^1, P^2, \cdots, P^b, \, w^1, w^2, \cdots, w^S], \\
[\ \underline{\hspace{0.3cm}}, \, W_i] &= \mathcal{L}_i([P_{i-1}, W_{i-1}]), \quad i = 1, \dots, t, \\
[P_j, \, W_j] &= \mathcal{L}_j([P_{j-1}, W_{j-1}]), \quad j = t+1, \dots, K \\
\mathbf{z} &= \texttt{TextProj} (w_K^S),
\end{align}
\noindent
where $\mathcal{L}_i$ denotes the $i$-th Transformer block, and $\texttt{TextProj}(\cdot)$ projects the final [EOS] token representation to a fixed embedding space shared with the image encoder. For BiomedCLIP, $N = 256$, and the final token $w_K^S$ corresponds to the [EOS] token. The prompt tokens are actively updated only for the first $t$ layers, which defines the depth of text prompting. The resulting text embedding $\mathbf{z} \in \mathbb{R}^{D}$ captures rich biomedical semantics.

Instead of geometric prompts, we use the final text embeddings $\mathbf{T} = [\mathbf{z}_1, \cdots, \mathbf{z}_C] \in \mathbb{R}^{C \times D}$ as inputs to the SAM prompt encoder:
\begin{equation}
\bm{P}_{\text{text}} = \bm{E}_{\text{prompt}}^{\text{text}}(\mathbf{T}) \in \mathbb{R}^{C \times d},
\end{equation}
where $\bm{E}_{\text{prompt}}^{\text{text}}$ is a lightweight adapter implemented as a two-layer MLP with GELU activation:
\begin{equation}
\bm{E}_{\text{prompt}}^{\text{text}}(\bm{z}_i) = \bm{W}_2 \cdot \text{GELU}(\bm{W}_1 \bm{z}_i + \bm{b}_1) + \bm{b}_2,
\end{equation}
with parameters $\bm{W}_1 \in \mathbb{R}^{m \times h}$, $\bm{W}_2 \in \mathbb{R}^{m \times m}$, $\bm{b}_1, \bm{b}_2 \in \mathbb{R}^{m}$, and dimensions $h=512$ and $m=256$. This transformation ensures compatibility between the BiomedCLIP text embeddings and the SAM prompt token space. Finally, these text-derived prompt tokens are concatenated with the pre-trained dense mask embeddings extracted from the image encoder, forming the complete set of input prompts to the SAM decoder:
\begin{equation}
\bm{P} = [\bm{P}_{\text{text}}; \bm{P}_{\text{dense}}]
\end{equation}
\noindent
This complete set of prompt tokens $\bm{P}$ are then passed to SAM's mask decoder $\mathcal{D}_{\text{mask}}$, which performs cross-attention between $\bm{P}$ and the image embeddings $\bm{F}$ to predict the final segmentation masks.

\subsection{Iterative Segmentation Refinement Strategy}
\label{subsec:refinement}
To improve segmentation precision in challenging regions, we apply a lightweight segmentation refinement step after the initial SAM mask prediction. We extract geometric cues, specifically the bounding box and centroid of the predicted region, to generate geometric prompts (box and point). These geometric prompts are concatenated with the text-derived prompt tokens and dense mask embeddings to form a hybrid prompting approach, combining semantic guidance with geometric specificity, enabling SAM to refine its predictions with minimal computational overhead. Here, we use two iterations of segmentation refinement in our experiments:
\begin{equation}
\bm{P}_{\text{post}} = [\bm{P}_{\text{text}}; \bm{P}_{\text{dense}}; \bm{P}_{\text{geometric}}]
\end{equation}

\begin{figure*}
    \centering
    \begin{center}
    \includegraphics[width=2\columnwidth]{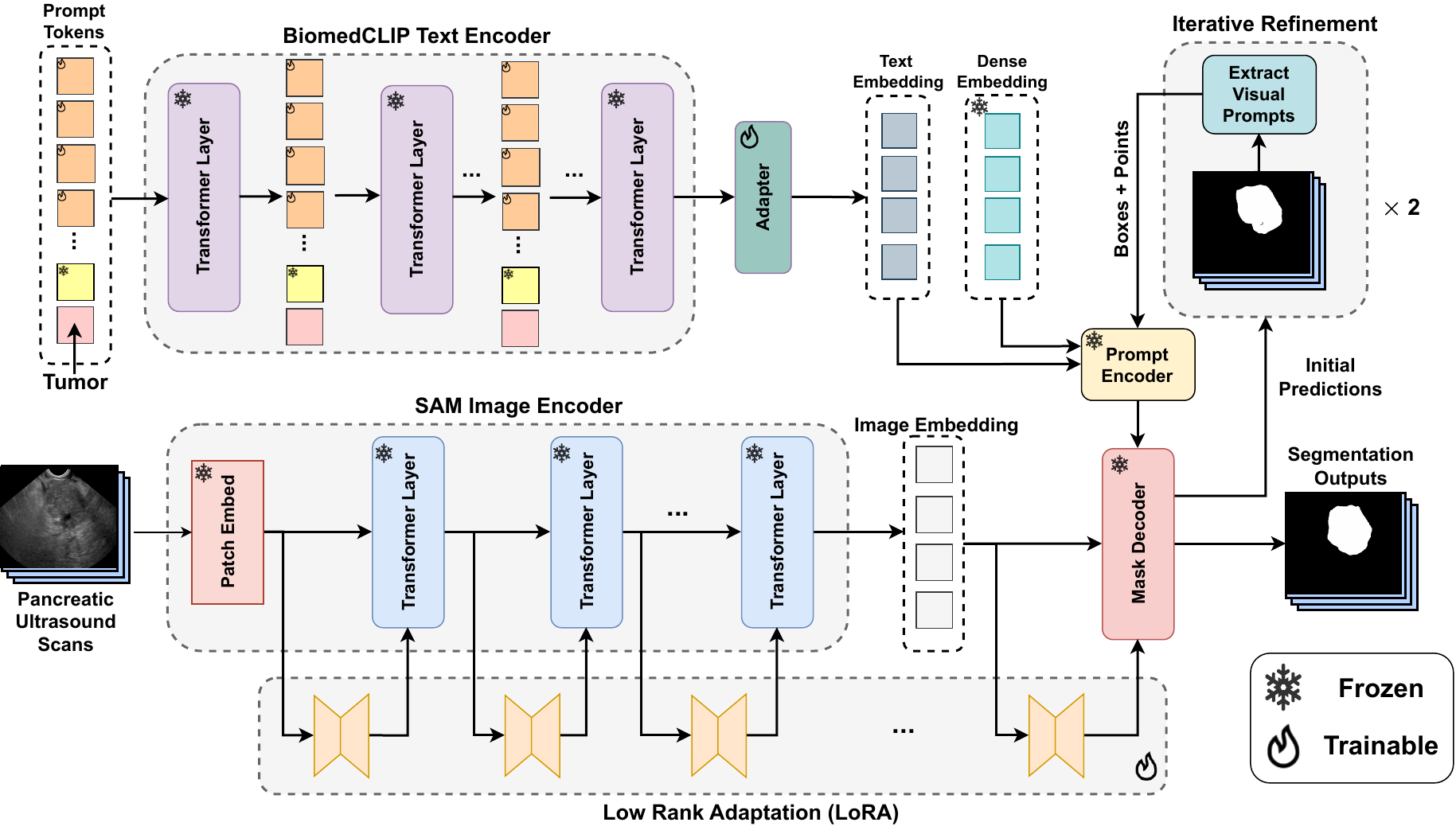}
    \end{center}
    \caption{Overview of the \textsc{TextSAM-EUS} framework. We finetune context tokens in the BiomedCLIP text encoder $\bm{E}_t$ to produce text embeddings $\mathbf{T}$, which are projected via an adapter $\bm{E}_{\text{prompt}}^{\text{text}}$ into the SAM prompt token space. These guide the SAM mask decoder $\mathcal{D}_{\text{mask}}$, which fuses them with image features $\bm{F} = \bm{E}_{\text{img}}(\bm{X})$ from the ViT-based image encoder $\bm{E}_{\text{img}}$, adapted using \textbf{LoRA}. An iterative segmentation refinement step incorporates geometric prompts (box and point) from initial predictions to enhance segmentation quality.}
    \label{fig:framework-fig}
\end{figure*}
\section{Experiments and Results}
\label{sec:results}

\subsection{Dataset Overview}
We evaluate our method using the publicly available Endoscopic Ultrasound Database of the Pancreas \cite{jaramillo2020endoscopic}, which contains high-resolution ultrasound frames acquired during EUS-guided procedures. The dataset includes labeled pancreatic tumor regions annotated by clinical experts, offering valuable pixel-wise ground truths for segmentation tasks. To ensure robust evaluation and avoid data leakage, we performed a strict patient-wise data split on all 18 patients in the dataset with pancreatic tumors. The final training set comprised 11,363 EUS images from 12 patients. The validation set contained 986 images from 2 additional patients, while the segmentation test set included 4,185 images from 4 held-out tumor patients. All images are grayscale at a resolution of 711 $\times$ 457 pixels.

\subsection{Baseline Models}
\label{subsec:baselines}

We compare \textsc{TextSAM-EUS} against three categories of DL methods, all trained and evaluated on the same pancreatic EUS data splits (Section~\ref{sec:results}). First, we train two popular fully supervised models, nnUNet and SwinUNet.  While nnUNet \cite{isensee2021nnu} automatically configures U-Net architectures and pre-/post-processing for optimal performance on each dataset, Swin-UNet \cite{cao2022swin} integrates a shifted-window Transformer encoder into a U-Net-like configuration to capture both global context and local detail. Second, we evaluate two non-SAM-based foundation models, including BiomedParse \cite{biomedparse}, a vision–language model pretrained on biomedical image–text pairs and adapted for prompt-based segmentation, and UniverSeg \cite{butoi2023universeguniversalmedicalimage}. Lastly, we benchmark SAM, and five additional SAM-based methods: SAMed \cite{samed}, AutoSAM \cite{autosam}, AutoSAMUS \cite{lin2024samus}, SPFS-SAM\cite{spfs}, and MedSAM \cite{medsam}, whose architectures are described in Section \ref{sec:related-work}. We fine-tuned SAMed, AutoSAM, AutoSAMUS, and SPFS-SAM on our training set, whereas we evaluated MedSAM and the original SAM using their pre-trained weights. For SAMUS, we report results using both the pre-trained and fine-tuned weights, since it is a tunable foundation model.

\subsection{Implementation Details}
\label{subsec:implementation-details}
For the baseline models expecting point prompts (SAMUS, SAM), one point was randomly sampled per ground truth. For models expecting bounding boxes (MedSAM, SAM), tight bounding boxes were extracted from ground truths, then perturbed 10-15 pixels per edge to simulate human-provided inputs. Since TextSAM-EUS’s refinement step uses both a point and a box, to test its performance with manual prompts, we passed it one ground-truth point and one perturbed bounding box for the manual-prompt variant. For BiomedParse \cite{biomedparse}, which requires a text prompt at inference time, we supplied the fixed text prompt ``\texttt{pancreatic tumor in ultrasound scan}'' for every image. For UniverSeg, we experimented with both 16-shot and 32-shot fine-tuning. All trained baseline models were optimized using the hyperparameter settings recommended by their original papers. Our proposed TextSAM-EUS model was trained on the training set for 5 epochs using a learning rate of 0.001 and a batch size of 1. We employ the AdamW optimizer \cite{loshchilov2017decoupled} with a weight decay of 0.01 and adopt a cosine annealing learning rate scheduler. The model selection is based on the lowest validation loss. We use a composite loss function consisting of \textbf{Dice loss} and \textbf{Binary Cross-entropy (BCE) loss}. The BiomedCLIP text encoder is configured with 4 context tokens and a depth of 12 layers (threaded through the first 12 Transformer layers of the prompt encoder). We use ``\texttt{tumor}'' as the class name, randomly initialize the context tokens (see Fig. \ref{fig:framework-fig}), and apply LoRA with a rank of 16 for parameter-efficient fine-tuning. We evaluate segmentation performance using the Dice Similarity Coefficient and the Normalized Surface Distance (computed at a boundary tolerance of 3 pixels), reporting results as mean ± standard deviation on a 0-100 \% scale across all test cases. The segmentation performances between methods are compared using two-tailed paired-sample \(t\)-tests, where a p-value $<$ 0.05 indicates a statistically significant difference. Training was done on a single NVIDIA A100 GPU (40GB RAM).

\subsection{Segmentation Results}
Table~\ref{tab:model-comparison} reports DSC and NSD on the held-out test set. Methods are grouped according to the baseline categories defined in Section~\ref{subsec:baselines}. \textsc{TextSAM-EUS} achieves the highest DSC (82.69\,\%) and NSD (85.28\,\%) among approaches requiring no manual prompts (p $<$ 0.05). Compared with the strongest automatic SAM variants (AutoSAM and AutoSAMUS), our model improves DSC by $\sim$ 1.4\% and 1.65\%, respectively, while tuning much fewer parameters (1.69 M vs.\ 41.56 M and 1.69 M vs.\ 8.86 M). MedSAM and the automatic \textsc{TextSAM-EUS} yield nearly identical DSC (82.69 vs.\ 82.66; \(p=0.906\)), while \textsc{MedSAM} shows a slight advantage in NSD (85.28 vs. 85.75; \(p=0.035\)). However, note that MedSAM is supplied with ground truth bounding boxes at inference. It is also important to note that our method significantly outperforms SAMed \cite{samed}, which also utilizes LoRA for adapting the image encoder and mask decoder modules, but only tunes the dense embedding, whereas our approach injects rich biomedical textual semantics to boost segmentation performance, highlighting the benefit of the deep text prompting. When supplied with additional manual geometric prompts, \textsc{TextSAM-EUS} attains a significantly higher DSC than MedSAM (\(p=0.039\)), although its difference in NSD is not significant (\(p=0.801\)), despite a marginally higher average NSD. These results demonstrate that our automatic text-driven \textsc{TextSAM-EUS} significantly outperforms all automatic baselines, and when supplied with manual prompts matches or exceeds other manual-prompted models. 

\begin{table}[ht]
\centering
\caption{Segmentation performance comparison across models. DSC and Normalized Surface Distance (NSD) metrics are reported as mean ± standard deviation. Models using ground truth prompts at inference represent upper bounds.}
\label{tab:model-comparison}
\tablestyle{-13pt}{1.1}
\arrayrulecolor{black}
\setlength\arrayrulewidth{1pt}
\addtolength{\tabcolsep}{+16pt}
\resizebox{\columnwidth}{!}{%
\begin{tabular}{lcc}
\toprule
\textbf{Model Name} & \textbf{DSC (\%) $\uparrow$} & \textbf{NSD (\%) $\uparrow$} \\
\midrule
\rowcolor{gray!15}
\multicolumn{3}{c}{UNet-based Approaches} \\
nnUNet                 & 79.94 ± 19.96  & 82.30 ± 20.24\\
SwinUNet               & 59.69 ± 23.33  & 10.22 ± 6.64
\\
\midrule
\rowcolor{gray!15}
\multicolumn{3}{c}{Foundation Models} \\
BiomedParse    & 37.00 ± 37.76  & 38.92 ± 36.81   \\
UniverSeg (16 shots)    & 42.92 ± 6.55  & 47.54 ± 6.86   \\
UniverSeg (32 shots)   & 43.61 ± 6.70  & 48.03 ± 7.00   \\
\midrule
\rowcolor{gray!15}
\multicolumn{3}{c}{SAM-based Approaches (Automatic Prompts)} \\
SAMed   & 78.00 ± 16.95  & 80.62 ± 17.00 \\
AutoSAM  & 81.26 ± 16.98  & 83.79 ± 17.19 \\
AutoSAMUS  & 81.04 ± 16.49  & 83.80 ± 16.61 \\
SPFS-SAM  & 60.64 ± 9.69  & 63.13 ± 9.82 \\
\rowcolor{SkyBlue!10}
\textbf{TextSAM-EUS (Ours)} & \textbf{82.69 ± 15.14}  & \textbf{85.28 ± 15.21} \\
\midrule
\rowcolor{gray!15}
\multicolumn{3}{c}{SAM-based Approaches with Point/Box Prompts (Upper Bounds)} \\
SAMUS    & 70.76 ± 16.60  & 73.86 ± 16.57   \\
SAMUS (Finetuned)     & 82.81 ± 12.56  & 85.50 ± 12.72   \\
MedSAM             & 82.66 ± 6.80  & 85.75 ±  6.28  \\
SAM (Point)               & 39.80 ± 22.78   & 41.54 ± 23.83 \\
SAM (Box)                & 78.15 ± 18.06  & 83.70 ± 18.06  \\
\rowcolor{SeaGreen!10}
\textbf{TextSAM-EUS (Ours)}              & \textbf{83.10 ± 14.13}  & \textbf{85.70 ± 14.12}  \\
\bottomrule
\end{tabular}
}
\end{table}

\subsection{Ablation Experiments}

To determine the contributions of each component in TextSAM‐EUS, we conducted four ablation studies on the following: (1) LoRA adaptation rank; (2) length of learnable context tokens; (3) depth at which these tokens are processed; and (4) choice of geometric prompts in the post‐processing stage. All results are reported on the held‐out segmentation test set (mean ± standard deviation).

\subsubsection{Effect of LoRA Rank}
To determine the adaptation capacity required for effective tumor segmentation, we varied the LoRA rank $r$ from 1 to 16 (see Table~\ref{tab:rank-ablation}). At $ r=1$, the model fails to learn meaningful features (44.00\% DSC), while $r=4$ recovers performance (81.33\% DSC, 83.19\% NSD). Further enlargement to $r = 16$ produces a smaller, consistent improvement (82.69\% DSC, 85.28\% NSD), suggesting that the adapter saturates once it captures the dominant appearance variations of pancreatic tumors.

\begin{table}[ht]
\centering
\caption{Ablation study on the effect of LoRA rank on segmentation performance. Higher ranks generally improve performance, with rank 16 achieving the best results.}
\label{tab:rank-ablation}
\tablestyle{-13pt}{1.1}
\arrayrulecolor{black}
\setlength\arrayrulewidth{1pt}
\addtolength{\tabcolsep}{+16pt}
\resizebox{\columnwidth}{!}{%
\begin{tabular}{cccc}
\toprule
LoRA Rank ($r$) & Params.\# & \textbf{DSC (\%) $\uparrow$} & \textbf{NSD (\%) $\uparrow$} \\
\midrule
1 & 1,043,812 & 44.00 ± 28.98 & 46.99 ± 29.34 \\
4 & 1,172,068 & 81.33 ± 16.53 & 83.19 ± 16.66 \\
\rowcolor{gray!20} 16 & 1,685,092 & \textbf{82.69 ± 15.14}  & \textbf{85.28 ± 15.21} \\
\bottomrule
\end{tabular}
}
\end{table}
\subsubsection{Effect of Context Token Length}
Table \ref{tab:length-ablation} examines the length $b$ of the learnable context prompt, demonstrating that four tokens provide the best DSC and NSD scores. Increasing to eight tokens introduces optimization instability and degrades accuracy, whereas 12 tokens partially recover performance but do not surpass the four-token setting. These observations indicate that a concise prompt encodes sufficient semantic detail for the target binary task.

\begin{table}[ht]
\centering
\caption{Ablation study on the effect of Number of Context Tokens on segmentation performance.}
\label{tab:length-ablation}
\tablestyle{-13pt}{1.1}
\arrayrulecolor{black}
\setlength\arrayrulewidth{1pt}
\addtolength{\tabcolsep}{+24pt}
\resizebox{\columnwidth}{!}{%
\begin{tabular}{cccc}
\toprule
Context Length ($b$) &  \textbf{DSC (\%) $\uparrow$} & \textbf{NSD (\%) $\uparrow$} \\
\midrule
\rowcolor{gray!20} 4 & \textbf{82.69 ± 15.14}  & \textbf{85.28 ± 15.21} \\
8 &  66.74 ± 21.68 & 68.29 ± 21.86 \\
 12 & 82.10 ± 15.20  & 83.99 ± 15.27 \\
\bottomrule
\end{tabular}
}
\end{table}
\begin{figure*}[ht]
    \centering
    \begin{center}
    \includegraphics[width=1.73\columnwidth]{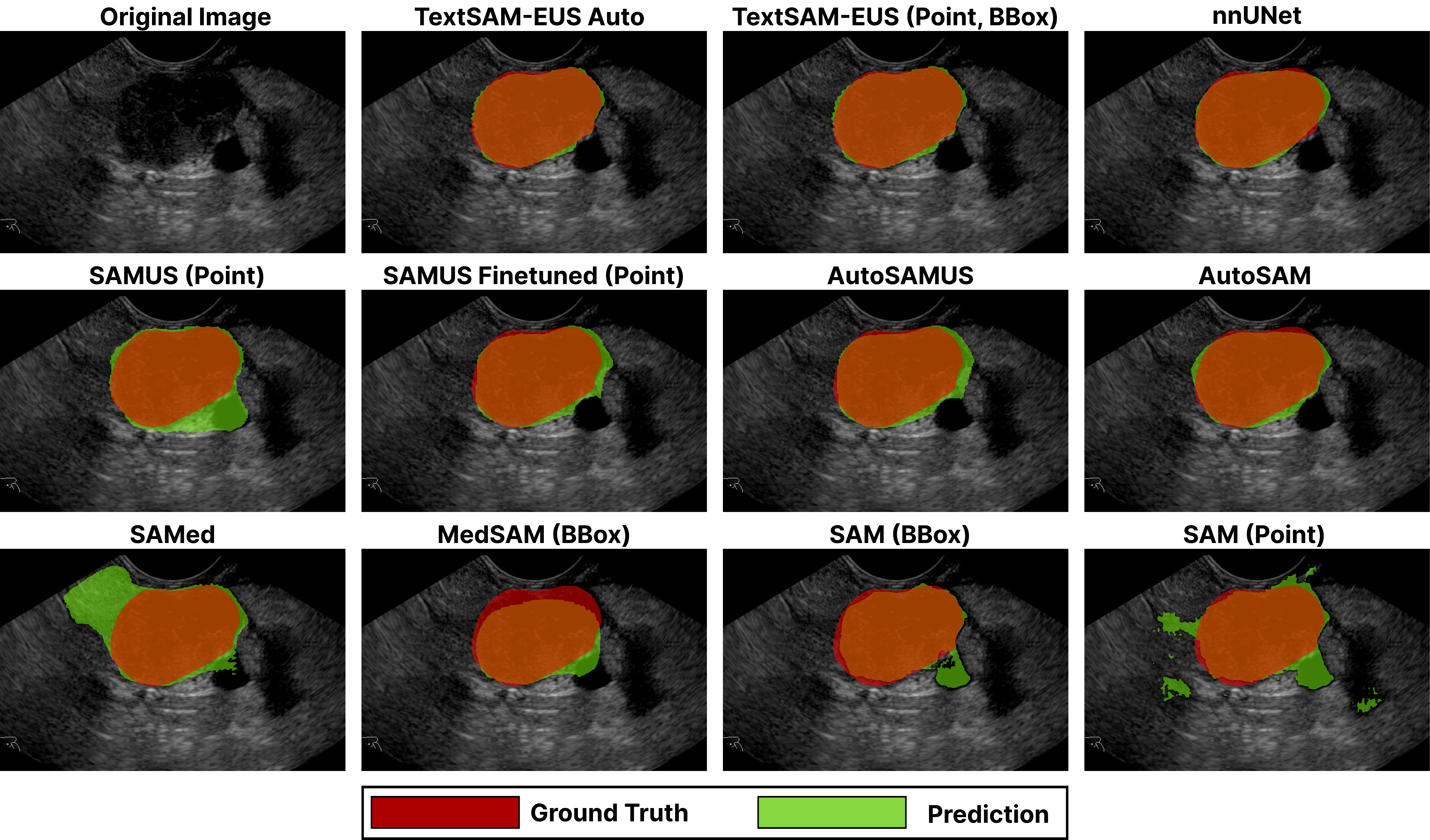}
    \end{center}
    \caption{Qualitative comparison on a representative EUS slice, comparing the most competitive nine baselines alongside \textsc{TextSAM-EUS} in its fully automatic and manual–prompt variants.  
    Green denotes the predicted mask, red the ground-truth, and orange the regions where the two overlap. Models supported with manual prompts have parentheses describing prompt types, and the remaining are automatic.}
    \label{fig:qualitative-fig}
\end{figure*}
\subsubsection{Effect of Prompt-injection depth}

The BiomedCLIP text encoder contains 12 Transformer blocks. Prompt-injection depth $t$ specifies how many of the early blocks (layers $1$ to $12$) receive the tunable deep prompt tokens, with the remaining layers running unchanged. Table \ref{tab:depth-ablation} shows that performance rises steadily with deeper processing and peaks when the prompts are threaded through all layers ($t = 12$), confirming that the context tokens must influence most of the encoder to guide segmentation effectively.

\begin{table}[ht]
\centering
\caption{Ablation study on the effect of learnable prompts depth on segmentation performance.}
\label{tab:depth-ablation}
\tablestyle{-13pt}{1.1}
\arrayrulecolor{black}
\setlength\arrayrulewidth{1pt}
\addtolength{\tabcolsep}{+24pt}
\resizebox{\columnwidth}{!}{%
\begin{tabular}{cccc}
\toprule
Layer Depth ($t$) &  \textbf{DSC (\%) $\uparrow$} & \textbf{NSD (\%) $\uparrow$} \\
\midrule
1 & 78.32 ± 19.05 & 80.97 ± 19.29 \\
4 &  76.22 ± 19.47 & 78.60 ± 19.67 \\
9 &  80.52 ± 17.71 & 82.36 ± 17.85 \\
\rowcolor{gray!20} 12 & \textbf{82.69 ± 15.14} & \textbf{85.28 ± 15.21} \\
\bottomrule
\end{tabular}
}
\end{table}

\subsubsection{Effect of Geometric Prompts in Refinement Stage}

Table \ref{tab:prompt-ablation} evaluates geometric cues that are automatically extracted from our text‐prompt predicted masks. Specifically, we compute the bounding box and the centroid of the initial mask output by TextSAM‐EUS and feed these geometry‐derived prompts back into the refinement stage. The first row represents the performance of the framework without the iterative segmentation refinement stage, i.e., using only the language-guided predicted mask. Even without geometric guidance, the language‐only mask achieves 82.00\% DSC. Using either the extracted bounding box or the centroid alone yields modest improvements. When both automatically‐derived box and centroid prompts are combined with the learned text prompts, we observe the highest DSC and NSD scores, and adding extra random points offers no further benefit.

\begin{table}[ht]
\centering
\caption{Effect of different geometric prompts in the Segmentation Refinement stage}
\label{tab:prompt-ablation}
\tablestyle{-13pt}{1.1}
\arrayrulecolor{black}
\setlength\arrayrulewidth{1pt}
\addtolength{\tabcolsep}{+17pt}
\resizebox{\columnwidth}{!}{%
\begin{tabular}{cccc}
\toprule
BBoxes & Points & \textbf{DSC (\%) $\uparrow$} & \textbf{NSD (\%) $\uparrow$} \\
\midrule
\rowcolor{gray!12} \xmark & \xmark & 82.00 ± 17.40 & 84.61 ± 17.53 \\
\cmark & 1 Random     & 82.63 ± 15.11 & 85.22 ± 15.21 \\
\xmark & 1 Random     & 82.50 ± 16.04  & 85.10 ± 16.14  \\
\cmark   & 5 Random & 82.65 ± 14.77 & 85.25 ± 14.87 \\
\rowcolor{gray!20} \cmark & Centroid & \textbf{82.69 ± 15.14}  & \textbf{85.28 ± 15.21} \\
\bottomrule
\end{tabular}
}
\end{table}

\subsubsection{Summary}
Across all studies, we observe that moderate adapter capacity ($r = 16$), a compact prompt ($b = 4$), deep prompt injection (\(t = 12\)), and a single centroid point + bounding box segmentation refinement strategy yields the best trade-off between parameter efficiency and segmentation accuracy.  

\subsection{Qualitative Results}
Figure \ref{fig:qualitative-fig} illustrates segmentation results for an EUS image. Both variants of \textsc{TextSAM-EUS} (automatic and manual) closely align with the ground truth. The fully supervised nnUNet slightly under- and over-segments along the mask edge, but performs only marginally worse than \textsc{TextSAM-EUS}. The recent SAM extensions, however, produce false positive regions (SAM-Point, SAMUS-Point, SAMed), often overshooting when relying on bounding boxes (SAM-BBox, MedSAM). AutoSAM and AutoSAMUS reduce manual effort but still leave small gaps or excess masks in this image example. 

\section{Discussion}
Our results confirm that adding linguistic context can guide segmentation as effectively as manually drawn geometric prompts in SAM, which requires radiological knowledge. We also demonstrate that text prompt learning outperforms other architectures used in SOTA SAM-based methods, manual and automatic alike. For example, SAMUS, which attaches a CNN to SAM’s ViT backbone and fine-tunes its 39.65 M parameters has worse performance than \textsc{TextSAM-EUS}, which only trains 1.69 M parameters.  Ultimately, our findings suggest that alignment between text tokens and the ViT feature grid is more valuable than adding an external CNN path for pancreatic cancer EUS, and possibly other domains as well. In a preliminary ablation experiment, swapping the SAMUS encoder into our framework yielded a maximum of 67.53\% DSC under LoRA fine-tuning, confirming the extra CNN did not help. Overall, our findings point to the strength of learnable text prompts in providing high-level semantic guidance directly to the ViT features, rather than relying on local texture filters from an added CNN branch. Beyond pancreatic tumors in EUS, our text-driven approach could potentially be applied more broadly, particularly when annotated medical datasets are limited, as the EUS dataset we used only included 18 patients. Interestingly, while manual geometric prompts can be omitted with learnable text prompts and fine-tuning, our iterative segmentation refinement block demonstrates that derived geometric prompts from text-based predictions still slightly improve mask quality. Furthermore, the low trainable parameter count indicates the model has the potential to be used in clinics with lower computational resources. In future works, we plan to extend our framework to multi-class segmentation, which may involve further adapting the prompt design. We also aim to evaluate the TextSAM framework on additional applications and imaging modalities.

\section{Conclusion}

In summary, we have introduced \textsc{TextSAM-EUS}, the first text-prompt-learning-based SAM for medical image segmentation. By integrating prompt learning with BiomedCLIP and LoRA-based tuning of SAM, our framework learns by tuning just 0.86\% of the total parameters (1.69 M), yet delivers accurate tumor masks without any manual geometric prompts. On the public Endoscopic Ultrasound Database of the Pancreas, \textsc{TextSAM-EUS} significantly outperformed all competing automatic methods from different categories in consideration. When supplied with manual prompts, it also matches or exceeds the best manual-prompt-guided baselines. These results highlight the value of language-driven prompting for ultrasound segmentation and open the door to broader biomedical applications of promptable foundation models.

\noindent \textbf{Acknowledgments}

\noindent
This research was supported by grants from the MEDTEQ+ consortium,   Alpha Tau Medical, and MITACS. It was conducted as part of the TransMedTech Institute’s activities, thanks, in part, to the financial support of the Apogee Canada Research Excellence Fund. Y.X. is supported by the Fond de la Recherche du Québec – Santé (FRQS-chercheur boursier Junior 1) and Parkinson Quebec \cite{xiao2024frq}. P.S. is supported by the Natural Sciences and Engineering Research Council of Canada (596537) and the Fonds de recherche du Québec – Nature et technologies (B1X-348625) \cite{spiegler2025frq}. T.K. is supported by Fonds de recherche du Québec – Nature et technologies (B2X-363874) \cite{koleilat2025frq}. C.S.M. is a consultant for Alpha Tau Medical. M.K-O. is supported by the Fond de la Recherche du Québec – Santé (FRQS-chercheur boursier Junior 2). 

{
    \small
    \bibliographystyle{ieeenat_fullname}
    \bibliography{main}
}


\end{document}